\documentclass[a4paper, 11pt]{article}
\usepackage[a4paper,top=2cm,bottom=2.5cm,left=1.5cm,right=1.5cm,marginparwidth=1cm]{geometry}
\usepackage{graphicx} 
\usepackage{times}
\usepackage{natbib}
\usepackage[colorlinks=True,citecolor=blue]{hyperref}
\usepackage{enumitem}
\usepackage{amsmath,amsfonts,bm}
\usepackage{xcolor}
\usepackage{titlesec}
\usepackage{cancel}
\usepackage{xfrac}

\titleformat*{\section}{\large\bfseries}

\usepackage{array}

\setitemize{leftmargin=0.5em,topsep=0.1em,parsep=0.5pt,label=\raisebox{0.25ex}{\tiny$\bullet$}}
\setlist[enumerate]{leftmargin=*, label= {\arabic*.}, itemsep=0.5em}

\newcolumntype{H}{>{\setbox0=\hbox\bgroup}c<{\egroup}@{}}

\newcommand{\R}{\mathbb{R}}

\newcommand{\xb}{\bm{x}}

\newcommand{\n}[1]{n_{#1}}

\definecolor{predcolor}{gray}{0.95}
\definecolor{scorecolor}{gray}{0.95}
\definecolor{riskcolor}{gray}{0.95}

\title{Preference Models assume Proportional Hazards of Utilities}
\author{Chirag Nagpal\\Meta Superintelligence Labs (MSL)}
\date{July 2025}
\begin{document}

\maketitle


\begin{abstract}
    Approaches for estimating preferences from human annotated data typically involves inducing a distribution over a ranked list of choices such as the Plackett-Luce model. Indeed, modern AI alignment tools such as Reward Modelling and Direct Preference Optimization are based on the statistical assumptions posed by the Plackett-Luce model. In this paper, I will connect the Plackett-Luce model to another  classical and well known statistical model, the Cox Proportional Hazards model and attempt to shed some light on the implications of the connection therein.
\end{abstract}

\section{Introduction}
Modelling of human preferences is an important step in modern post-training pipelines for AI alignment. One popular approach of building such models of human preference is assuming that human preference rankings assume a Plackett-Luce \citep{plackett1975analysis, luce1959individual} distribution. In this monograph, I draw a somewhat remarkable connection of the popular statistical model for estimating lifetimes, the Cox Proportional Hazard model \citep{cox1972regression} to the Plackett-Luce model and then consequently to algorithms such as Direct Preference Optimization, a popular algorithm for aligning modern Artifical Intelligence \citep{ouyang2022training}. 

To the best of my knowledge, at the time of writing the connection between the Proportional Hazards model and the Plackett-Luce is relatively little known, and the subsequent connections to the AI alignment algorithms such as `\textit{Direct Preference Optimization}' \citep{rafailov2023direct} are not well appreciated. I believe that explcitly stating this connection will help the AI research community build on existing research in semi-parametric statistics to build better models of human preference.

\section{The Plackett-Luce and Bradley-Terry Model}

The Plackett-Luce model was independently introduced by Plackett and Luce as a probabilistic framework to model ranked data. Consider a list of $n$ observations $\{ ( \bm{x}_i, \bm{r}_i )_{i=1}^{n} \}$ where each example $i$ consists of features $\xb_i$ and  $r_i \in [n]$ represents its observed ranking or position in the data. The central idea behind the Plackett-Luce model is to express the probability of observing a particular ranking of these items as a function of the features associated with each item. 

Specifically, suppose $\sigma(\cdot)$ is a permutation on the set $[n]$ that defines the order of the items such that $\sigma(i)$ is the index of the item with the $i^\textrm{th}$ rank. The Plackett-Luce model assigns a probability to the event that the ranking is exactly $\{ \sigma(1) \succ ... \succ \sigma(n) \}$ based on their relative scores. Formally, this probability is given by the product:%
\begin{align}
\small
 \mathbf{P}\big(\sigma(1) \succ ... \succ \sigma(n) | \bm{f}, \{ \bm{x}_i\}_{i=1}^{n}\big) = \prod_{i=1}^{n} \frac{\exp\bigg(f(\bm{x}_{\sigma(i)})\bigg)}{\sum\limits_{j=i}^{n} \exp\bigg(f(\bm{x}_{\sigma(j)})\bigg) }.
 \label{eqn:pl}
\end{align}
 \noindent Here $\sigma(j)$ represents the $j^\text{th}$ ranked item in the datum. $\bm{f}: \mathbb{R}^d \to \mathbb{R}$ is some function that operates on the covariates and scores the item. I use the terms Bradley-Terry and Plackett-Luce interchangeably, since the Bradley-Terry model is a special case of the model above when the number of choices $(k=2)$. 
 
 Replacing $\bm{f}$ with $\log \frac{\pi(\cdot)}{\pi^{\text{ref}}(\cdot)}$ in Equation \ref{eqn:pl} recovers the popular AI alignment algorithm Direct Preference Optimization (DPO) \citep{rafailov2023direct}. In the case of DPO $\pi(\cdot | \texttt{Input})$ refers to the probability of a response given an input to a language model.

\newpage

\begin{figure}[!htbp]
\begin{minipage}{0.5\textwidth}
\centering
\textbf{ a) Cumulative Density Function}
\end{minipage}%
\begin{minipage}{0.5\textwidth}
\centering
\textbf{b) Hazard Function}
\end{minipage}\\
\begin{minipage}{0.5\textwidth}
\centering
\includegraphics[width=0.9\textwidth]{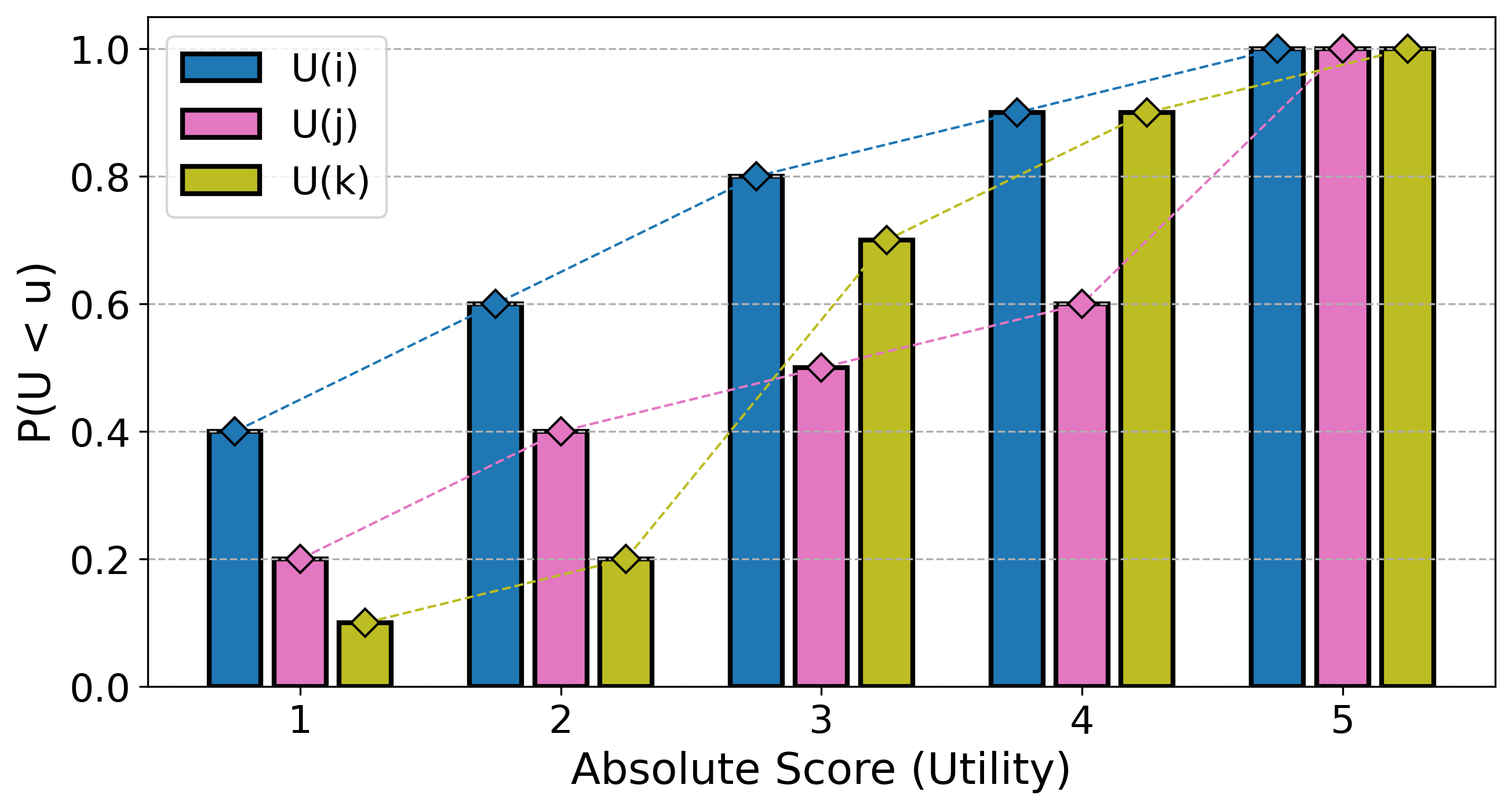}
\end{minipage}\hfill
\begin{minipage}{0.5\textwidth}
\centering
\includegraphics[width=0.9\textwidth]{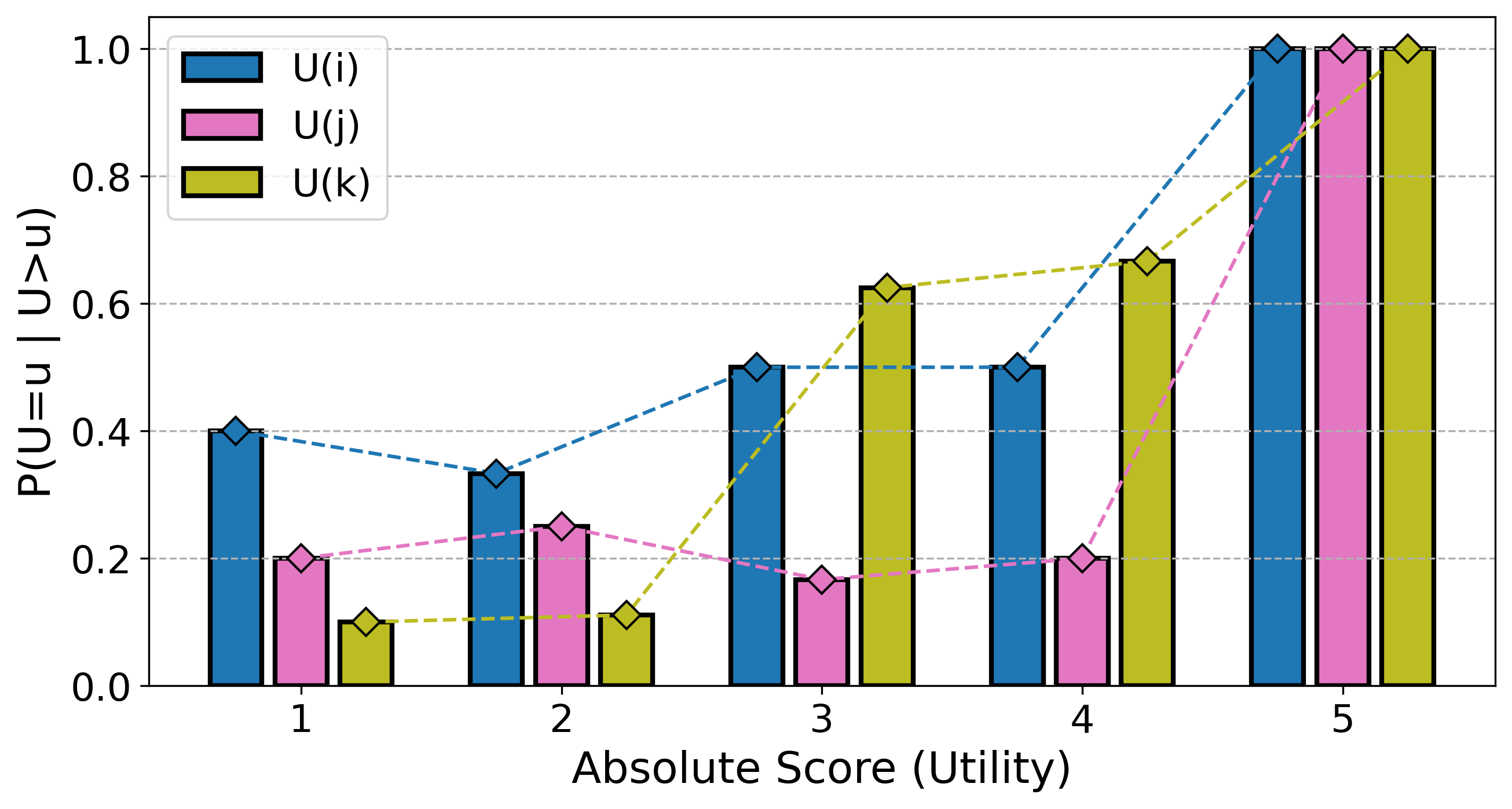}
\end{minipage}%

\caption{\textbf{The violation of the Proportional Hazards assumption in the Likert scale}. The cdfs of the utilities stochastically dominate each other ie. $\mathrm{Pr}(U(i)<u) > \mathrm{Pr}(U(j)<u), \, \forall u \in \mathcal{U}$; while this condition is violated for $U(k)$. \textbf{Plackett-Luce can recover preferences corresponding to} $(i, j)$ \textbf{but not when PH is violated as in} $(i, k)$. }
\label{fig:ph_violation}
\end{figure}
\section{The Cox Proportional Hazards (PH) Model}

The Cox model is popular in bio-statistics, reliability engineerring and acturial sciences for the estimation of \textit{time-to-event} outcomes. With slight modification instead of modeling times let us consider a dataset of $\n{}$ examples $\{ (\xb_i, u_i)\}_{i=1}^n$. Each example $i$ consists of features $\xb_i  \in \R^{d}$, a scalar \emph{utility} $u_i \in \mathbb{R}_+$. In the human preference setting utility is akin to a 5-point rating scale or time for a snippet of AI generated code to run. 

Given such a dataset, the Cox model's goal is to estimate a \emph{hazard rate} for the utilities $u$ conditioned on the features $\xb$. Here, the hazard rate is a function that describes the instantaneous rate of change of the utility function at a certain value of utility. More formally the hazard rate is:
\begin{align}
\small
\bm{\lambda}(u) := \mathop{\lim}_{\Delta  u\to 0} \frac{\mathbf{P}({ u<U \leq u+ \Delta u \mid U > u})}{\Delta u}.
\end{align}

Estimation in terms of the hazard rate $\bm{\lambda}(u)$ is natural in survival analysis and reliability engineering and can be used to estimate other quantities of interest  such as the survival function.\footnote{The survival rate is the negative exponent of the cumulative hazard, ie. $\bm{S}(u) = \exp \big( - \int_0^{u} \bm{\lambda}(u) \big).$} This model assumes that the ratio between the hazard rate for a point with features $\xb$ at a utility level $u$ changes with respect to the \emph{baseline hazard rate} $\bm \lambda_0(u)$\footnote{The base hazard rate $\bm{\lambda}$ is an infinite dimensional functional parameter of the model, which is estimated non-parametrically.} and that the rate of change is determined by the some function ${f}$ operating on the covariates $\bm{x}$. We assume that the conditional hazard rate function follows a \emph{proportional hazards} (PH) model:
\begin{align}
\small
\bm \lambda(U= u \mid X=\xb) := \bm \lambda_0(u)  \exp{\big(f(\bm{x})\big)}.
\label{eqn:PH}
\end{align}

 \noindent The \textit{proportionality of hazards} follows naturally from Equation \ref{eqn:PH} above since it implies that for any $(i, j)$ pair:  $$\forall u \in \mathcal{U}, \quad {\LARGE{ \sfrac{\bm{\lambda}(u | X= \bm{x}_i)}{\Large \bm{\lambda}(u| X= \bm{x}_j)}}} = \mathrm{constant} \,.$$  

The parameters of the Cox Proportional Hazards model $\bm{f}$ are estimated by minimizing the \textit{partial likelihood}. Given a dataset we define the partial likelihood $\mathcal{L}(\bm f)$ imposed by the Cox Proportional Hazards model as:
\begin{align}
\small
\mathcal{L}(f) = 
 \prod_{i = 1}^{n} \frac{ \bcancel{ \bm\lambda_0 (u) } \exp \big( f(\bm x_i) \big)}{\sum\limits_{j: \, u_j \geq u_i } \bcancel{ \bm\lambda_0 (u) }  \exp \big( f(\bm x_j) \big)}  =  \prod_{i = 1}^{n} \frac{ \exp \big( f(\bm x_i) \big)}{\sum\limits_{j: \, u_j \geq u_i} \exp \big( f(\bm x_j) \big)}.
 \label{eqn:partial}
\end{align}

\noindent The partial likelihood is called such as it is independent of $\bm{\lambda}_0 (\cdot)$ representing the base hazard. Typically $\bm{\lambda}_0 (\cdot)$ is treated a \textit{nuisance} parameter and not estimated. One can however, estimate the cumulative density of the utilities using the following non-parametric estimator \citep{breslow1972}:
\begin{align}
\widehat{\bm{S}}_0(u) = \exp \left( - \sum\limits_{i:\, u_i<u}\frac{1}{\sum \limits_ {j:\, u_j \geq u_i}  \exp \bigg(\widehat{\bm{f}}(\bm x_j ) \bigg)} \right) \quad \text{and} \quad  \widehat{\mathbf{P}}(U<u|X=\bm{x}, \widehat{\bm{f}} ) = 1 - \widehat{\bm{S}}_0(u) ^{\exp\big(\widehat{\bm{f}}(x)\big)} .
\end{align}

\newpage

\section{The Connection and implication for AI Alignment}



When typically working with Plackett-Luce models we only observe the relative ranking order. Let us now introduce some explicit notion of utility $\mathcal{U}(\cdot)$, that determines the intrinsic (absolute) quality or value of an associated choice $i \succ j \iff \mathcal{U}(i) > \mathcal{U}(j).$ Comparing Equation \ref{eqn:partial} to Equation \ref{eqn:pl} should make the connection abundantly clear. Assuming the rankings are associated with an observed scalar utility $u$, somewhat remarkably \textbf{the Plackett-Luce assumptions recovers exactly the same model as the Cox PH model.}\\

 \noindent \textbf{Plackett-Luce and its derivatives are sensitive to the assumption of \textit{Proportional Hazards.}} Models based on the Plackett-Luce assumptions (such as Bradley-Terry Reward Models or Direct Preference Optimization) are restricted to modelling preferences whose underlying utility functions are stochastically dominated (Figure  \ref{fig:ph_violation}). When fitted to data arising from utilities that violate PH, the Plackett-Luce model likely will mis-estimate human preferences.\\
 \noindent \textbf{This is more than just a theoretical insight and has practical consequences}. \textbf{Real world annotations on polarizing concepts manifest population level heterogeneity, in such situations preferences will likely be mis-estimated.}\\
    
 \noindent \textbf{Estimating the conditional distribution of utilties with preference data.} I mentioned that in the Cox model $\bm{\lambda}_0 (\cdot)$ is effectively a nuisance parameter that is not involved directly in the estimation. Now let us consider situations in the presence of data representing absolute utilities, such as point-wise feedback, the baseline hazard can simultaneously be estimated non-parametrically \citep{breslow1972, lin2007breslow}. This can be used to recover an estimate of the true distribution over the absolute value of utility, leading to better estimation of relative utility.


\section{Conclusion}

In this paper, I have attempted to bring about a relatively little known connection between Plackett and Luce's statistical model of choice and the Cox's classic Proportional Hazards model. Although some work \citep{chen2024preference, maystre2022temporally} has similar flavor, they do not explicitly mention  this connection\footnote{Although I have not had a correspondence with either, I am inclined to believe that \citeauthor{ranganath2016deep}~and \citeauthor{maystre2019efficient}~maybe privy to the connection with `\textit{Proportional Hazards}' due to their prior art in the area.}. I have further shed some light on the implications for the design of better approaches to align AI with human preferences.

\small 
\bibliography{ref}

\begin{thebibliography}{}

\bibitem[Breslow, 1972]{breslow1972}
Breslow, N.~E. (1972).
\newblock Discussion of the paper by {D}.{R}.\ {C}ox.
\newblock {\em J R Statist Soc B}.

\bibitem[Chen et~al., 2024]{chen2024preference}
Chen, A., Malladi, S., Zhang, L., Chen, X., Zhang, Q.~R., Ranganath, R., and Cho, K. (2024).
\newblock Preference learning algorithms do not learn preference rankings.
\newblock {\em Advances in Neural Information Processing Systems}.

\bibitem[Cox, 1972]{cox1972regression}
Cox, D.~R. (1972).
\newblock Regression models and life-tables.
\newblock {\em Journal of the Royal Statistical Society: Series B (Methodological)}.

\bibitem[Lin, 2007]{lin2007breslow}
Lin, D. (2007).
\newblock On the breslow estimator.
\newblock {\em Lifetime data analysis}, 13(4).

\bibitem[Luce et~al., 1959]{luce1959individual}
Luce, R.~D. et~al. (1959).
\newblock {\em Individual choice behavior}, volume~4.
\newblock Wiley New York.

\bibitem[Maystre, 2019]{maystre2019efficient}
Maystre, L. (2019).
\newblock {\em Efficient learning from comparisons}.
\newblock Gesellschaft f{\"u}r Informatik eV.

\bibitem[Maystre and Russo, 2022]{maystre2022temporally}
Maystre, L. and Russo, D. (2022).
\newblock Temporally-consistent survival analysis.
\newblock {\em Advances in Neural Information Processing Systems}, 35:10671--10683.

\bibitem[Ouyang et~al., 2022]{ouyang2022training}
Ouyang, L., Wu, J., Jiang, X., Almeida, D., Wainwright, C., Mishkin, P., Zhang, C., Agarwal, S., Slama, K., Ray, A., et~al. (2022).
\newblock Training language models to follow instructions with human feedback.
\newblock {\em Advances in neural information processing systems}, 35.

\bibitem[Plackett, 1975]{plackett1975analysis}
Plackett, R.~L. (1975).
\newblock The analysis of permutations.
\newblock {\em Journal of the Royal Statistical Society Series C: Applied Statistics}.

\bibitem[Rafailov et~al., 2023]{rafailov2023direct}
Rafailov, R., Sharma, A., Mitchell, E., Manning, C.~D., Ermon, S., and Finn, C. (2023).
\newblock Direct preference optimization: Your language model is secretly a reward model.
\newblock {\em Advances in neural information processing systems}.

\bibitem[Ranganath et~al., 2016]{ranganath2016deep}
Ranganath, R., Perotte, A., Elhadad, N., and Blei, D. (2016).
\newblock Deep survival analysis.
\newblock In {\em Machine Learning for Healthcare Conference}. PMLR.

\end{thebibliography}
\bibliographystyle{apalike}

\end{document}